\documentclass[10pt,a4paper,twocolumn,twoside]{IEEEtran}

\usepackage[T1]{fontenc}
\usepackage[latin9]{inputenc}
\usepackage{epsfig,psfrag}
\usepackage{amsmath,amsfonts,amssymb,amsxtra,bm}
\usepackage{graphicx}
\usepackage{color}
\usepackage{cite}

\newcommand{\be}{\begin{equation}}
\newcommand{\ee}{\end{equation}}
\newcommand{\ist}{\hspace*{.3mm}}
\newcommand{\rmv}{\hspace*{-.3mm}}
\newcommand{\iist}{\hspace*{1mm}}

\allowdisplaybreaks

\begin{document}
\title{Sigma Point Belief Propagation
\vspace{1mm}
}
\author{Florian Meyer, \emph{Student Member, IEEE}, Ondrej Hlinka, \emph{Member, IEEE}, and Franz Hlawatsch, \emph{Fellow, IEEE} 
\thanks{F.\ Meyer, O.\ Hlinka, and F.\ Hlawatsch
are with the Institute of Telecommunications, Vienna University of Technology, A-1040 Vienna, Austria (email: \{florian.meyer,$\ist$ondrej.hlinka,$\ist$franz.hlawatsch\}@tuwien.ac.at).
This work was supported by FWF grant S10603 and by the Newcom\# Network of Excellence in Wireless
Communications of the European Commission.}}
\maketitle

\begin{abstract}
The sigma point (SP) filter, also known as unscented Kalman filter, is an attractive alternative to the extended Kalman filter and the particle filter.
Here, we extend the SP filter to nonsequential Bayesian inference corresponding to loopy factor graphs.
We propose \emph{sigma point belief propagation} (SPBP) as a low-complexity approximation of the belief propagation (BP) message passing scheme. 
SPBP achieves approximate marginalizations of posterior distributions  corresponding to (generally) loopy factor graphs. It is well suited for 
decentralized inference because of its low communication requirements. For a decentralized, dynamic sensor localization problem, 
we demonstrate that SPBP can outperform nonparametric (particle-based) BP while requiring significantly less computations and communications.
\end{abstract}


%%%%%%%%%%%%%%%%%%%%%%%%%%%%%%%%%%%%%%
\section{Introduction}
\label{sec:Intro}
%%%%%%%%%%%%%%%%%%%%%%%%%%%%%%%%%%%%%%SP

The sigma point (SP) filter, also known as unscented Kalman filter, is a sequential Bayesian estimator for nonlinear systems
that outperforms the extended Kalman filter while being typically less complex than the particle filter \cite{julier97anew,haykin2001ch7}.
Sequential Bayesian estimation corresponds to a ``sequential'' factor structure of the joint posterior probability density function (pdf).
For more general---possibly loopy---factor structures, the belief propagation (BP) message passing scheme can be used to (approximately) 
perform the marginalizations required for Bayesian inference \cite{loeliger}. Gaussian BP (GBP) \cite{weiss01} and nonparametric BP (NBP) \cite{ihler} 
are reduced-complexity approximations of BP that extend the Kalman filter \cite{haykin2001ch1} and the particle filter \cite{doucet}, respectively to 
general factor structures. GBP assumes a linear, Gaussian system and uses Gaussian message representations, whereas NBP is suited to nonlinear, 
non-Gaussian systems due to its use of particle representations.

Here, we propose the \emph{sigma point BP} (SPBP) message passing scheme as a new low-complexity approximation of BP 
for general nonlinear systems. SPBP extends the SP filter to general factor structures.  We demonstrate that the performance of SPBP can be similar
to or even better than that of NBP, at a far lower complexity. SPBP is well suited to certain distributed (decentralized) inference
problems in wireless sensor networks because of its low communication requirements---only a mean vector and a covariance matrix 
are communicated between neighboring sensors. We note that SPBP is different from the algorithm proposed in \cite{sun07},
which uses SPs to approximate BP in Bayesian networks and an information fusion technique to multiply messages. In contrast, SPBP is based 
on a factor graph and a reformulation of BP in higher-dimensional spaces. In addition to the advantages of factor graphs 
\pagebreak %%%%%%
discussed in \cite{loeliger}, this simplifies a decentralized implementation.

This letter is organized as follows. Some basics of SPs are reviewed in Section \ref{sec:SPbasics}. In Section \ref{sec:BP}, the system model 
is described and BP is reviewed. The SPBP scheme is developed in Section \ref{sec:SPBP}, and its computation and communication requirements 
are discussed in Section \ref{sec:compcom}. Section \ref{sec:simres} presents simulation results for a decentralized, cooperative, dynamic self-localization problem.

\vspace{-.5mm}

%%%%%%%%%%%%%%%%%%%%%%%%%%%%%%%%%%%%%%
\section{Sigma Point Basics}
\label{sec:SPbasics}
%%%%%%%%%%%%%%%%%%%%%%%%%%%%%%%%%%%%%%

\vspace{.5mm}

Consider a general (non-Gaussian) random vector $\mathbf{x} \rmv\in\rmv \mathbb{R}^J$ whose mean $\bm{\mu}_{\mathbf{x}} = \mathsf{E}\{\mathbf{x}\}$ 
and covariance matrix $\mathbf{C}_{\mathbf{x}} = \mathsf{E}\{ (\mathbf{x} \!-\! \bm{\mu}_{\mathbf{x}}) (\mathbf{x} \!-\! \bm{\mu}_{\mathbf{x}})^{\text{T}} \}$ are known,
and a transformed random vector $\mathbf{y} \!=\! H(\mathbf{x})$, where $H(\cdot)$ is a generally nonlinear function. SPs $\big\{ \mathbf{x}^{(j)} \big\}_{j=0}^{2J}$ 
and corresponding weights $\big\{ w_{\text{m}}^{(j)} \big\}_{j=0}^{2J}$ and $\big\{ w_{\text{c}}^{(j)} \big\}_{j=0}^{2J}$ are chosen such that the 
weighted sample mean $\tilde{\bm{\mu}}_{\mathbf{x}} = \sum_{j=0}^{2J} w_{\text{m}}^{(j)} \mathbf{x}^{(j)}$ and weighted sample covariance matrix
$\tilde{\mathbf{C}}_{\mathbf{x}} =  \sum_{j=0}^{2J} w_{\text{c}}^{(j)} (\mathbf{x}^{(j)} \!-\rmv \tilde{\bm{\mu}}_{\mathbf{x}})(\mathbf{x}^{(j)} \!-\rmv \tilde{\bm{\mu}}_{\mathbf{x}})^{\text{T}}$ 
are exactly equal to $\bm{\mu}_{\mathbf{x}}$ and $\mathbf{C}_{\mathbf{x}}$, respectively. Closed-form expressions of the SPs and weights are provided in 
\cite{haykin2001ch7}. The spread of the SPs around the mean $\bm{\mu}_{\mathbf{x}}$ can be adjusted via tuning parameters, whose choice depends on the 
dimension $J$ of $\mathbf{x}$ \cite{julier97anew,haykin2001ch7}. Next, each SP is propagated through $H(\cdot)$, resulting in 
$\mathbf{y}^{(j)} \!=H\big(\mathbf{x}^{(j)}\big)$, $j \rmv\in\rmv \{0,\ldots,2J\}$ (``unscented transformation'').
The\linebreak %%%%%% 
set $\big\{ \rmv\big( \mathbf{x}^{(j)}\rmv, \mathbf{y}^{(j)}\rmv, w_{\text{m}}^{(j)}\!,w_{\text{c}}^{(j)} \big)\rmv \big\}_{j=0}^{2J}$\!
then represents the joint second-order statistics of $\mathbf{x}$ and $\mathbf{y}$ in an approximate manner. In particular, $\bm{\mu}_{\mathbf{y}}$,  
$\mathbf{C}_{\mathbf{y}}$, and $\mathbf{C}_{\mathbf{xy}} = \mathsf{E}\{ (\mathbf{x} \!-\! \bm{\mu}_{\mathbf{x}}) (\mathbf{y} \!-\! \bm{\mu}_{\mathbf{y}})^{\text{T}} \}$ 
are approximated 
\vspace{-.5mm}
by
\begin{align}
\tilde{\bm{\mu}}_{\mathbf{y}} &= \sum_{j=0}^{2J} w_{\text{m}}^{(j)} \mathbf{y}^{(j)} \label{eq:MUandCOV_my}\\[.5mm]
\tilde{\mathbf{C}}_{\mathbf{y}} &= \sum_{j=0}^{2J} w_{\text{c}}^{(j)} (\mathbf{y}^{(j)} \!-\rmv \tilde{\bm{\mu}}_{\mathbf{y}})
  (\mathbf{y}^{(j)} \!-\rmv \tilde{\bm{\mu}}_{\mathbf{y}})^{\text{T}} \label{eq:MUandCOV_cy}\\[.5mm]
\tilde{\mathbf{C}}_{\mathbf{xy}} &= \sum_{j=0}^{2J} w_{\text{c}}^{(j)} (\mathbf{x}^{(j)} \!-\rmv \tilde{\bm{\mu}}_{\mathbf{x}})
  (\mathbf{y}^{(j)} \!-\rmv \tilde{\bm{\mu}}_{\mathbf{y}})^{\text{T}} . \label{eq:MUandCOV_cxy}
\end{align}
It has been shown in \cite{julier97anew} and \cite{haykin2001ch7} that these approximations are at least as accurate as those
resulting from a linearization (first-order Taylor series approximation) of $H(\cdot)$. Note also that the number $2J+1$ of SPs grows 
linearly with the dimension of $\mathbf{x}$ and is typically much smaller than the number of random samples in a particle representation.

Next, we consider the use of SPs for Bayesian estimation of a random vector $\mathbf{x}$ from an observed vector 
\[
\mathbf{z} \ist=\ist \mathbf{y} + \mathbf{n} \,, \quad \text{with} \;\, \mathbf{y} = H(\mathbf{x}) \,.
\]
Here, the noise $\mathbf{n}$ is statistically independent of $\mathbf{x}$ and generally non-Gaussian, with zero mean and known covariance matrix 
$\mathbf{C}_{\mathbf{n}}$. Bayesian estimation relies on the posterior pdf 
\begin{equation}
\label{eq:BayesianRule}
f(\mathbf{x}|\mathbf{z}) \ist\propto\ist f(\mathbf{z}|\mathbf{x})f(\mathbf{x}) \,,
\end{equation}
where $f(\mathbf{z}|\mathbf{x})$ is the likelihood function and $f(\mathbf{x})$ is the prior pdf.
Direct calculation of \eqref{eq:BayesianRule} is usually infeasible. An important exception is the case where $\mathbf{x}$ and $\mathbf{n}$ 
are Gaussian random vectors and $H(\mathbf{x}) = \mathbf{H} \mathbf{x}$ with some known matrix $\mathbf{H}$. Then 
$f(\mathbf{x}|\mathbf{z})$ is also Gaussian, and the posterior mean $\bm{\mu}_{\mathbf{x}|\mathbf{z}}$ and posterior covariance matrix 
$\mathbf{C}_{\mathbf{x}|\mathbf{z}}$ can be calculated as
\be
\hspace*{-1mm}\bm{\mu}_{\mathbf{x}|\mathbf{z}} =\ist \bm{\mu}_{\mathbf{x}} + \mathbf{K}(\mathbf{z} \rmv-\rmv \bm{\mu}_{\mathbf{y}}) \,, \quad\!\!
\mathbf{C}_{\mathbf{x}|\mathbf{z}} =\ist \mathbf{C}_{\mathbf{x}} - \mathbf{K} (\mathbf{C}_{\mathbf{y}} \rmv+\rmv \mathbf{C}_{\mathbf{n}}) \mathbf{K}^{\text{T}} \rmv, 
\label{eq:MessUpdate}
\vspace{.5mm}
\ee
where
\vspace{-1.3mm}
\be
\bm{\mu}_{\mathbf{y}} = \mathbf{H} \bm{\mu}_{x} \,, \quad
\mathbf{C}_{\mathbf{y}} = \mathbf{H} \mathbf{C}_{\mathbf{x}} \mathbf{H}^{\text{T}} 
\label{eq:mc}
\vspace{-2.5mm}
\ee
and 
\vspace{-1.3mm}
\be
\mathbf{K} \ist=\ist
\mathbf{C}_{\mathbf{x}\mathbf{y}}(\mathbf{C}_{\mathbf{y}} \rmv+\rmv \mathbf{C}_{\mathbf{n}})^{-1} , \quad
  \text{with} \;\, \mathbf{C}_{\mathbf{x}\mathbf{y}} = \mathbf{C}_{\mathbf{x}} \mathbf{H}^{\text{T}} .
\label{eq:K}
\ee
These expressions are used in the measurement update step of the Kalman filter \cite{haykin2001ch1}. The minimum mean-square error 
estimate of $\mathbf{x}$ is given by $\bm{\mu}_{\mathbf{x}|\mathbf{z}}$, and a characterization of the accuracy of estimation by $\mathbf{C}_{\mathbf{x}|\mathbf{z}}$.

In the general case of nonlinear $H(\cdot)$, the basic approximation underlying the extended Kalman filter \cite{haykin2001ch1} is obtained by 
using (essentially) \eqref{eq:MessUpdate}--\eqref{eq:K} with $\mathbf{H}$ being the Jacobian matrix resulting from a linearization of $H(\cdot)$. 
A more accurate alternative is to approximate $\bm{\mu}_{\mathbf{x}|\mathbf{z}}$ and $\mathbf{C}_{\mathbf{x}|\mathbf{z}}$ by means of SPs. For this,
we use \eqref{eq:MessUpdate} and the first equation in \eqref{eq:K}, with $\bm{\mu}_{\mathbf{y}}$, $\mathbf{C}_{\mathbf{y}}$, 
and $\mathbf{C}_{\mathbf{xy}}$ replaced by the SP approximations $\tilde{\bm{\mu}}_{\mathbf{y}}$, $\tilde{\mathbf{C}}_{\mathbf{y}}$, 
and $\tilde{\mathbf{C}}_{\mathbf{xy}}$ in \eqref{eq:MUandCOV_my}--\eqref{eq:MUandCOV_cxy}. This 
\vspace*{.7mm}
gives
\be
\hspace*{-1.5mm}\tilde{\bm{\mu}}_{\mathbf{x}|\mathbf{z}} =\ist \bm{\mu}_{\mathbf{x}} + \tilde{\mathbf{K}}(\mathbf{z} \rmv-\rmv \tilde{\bm{\mu}}_{\mathbf{y}}) \,, \quad\!\!
\tilde{\mathbf{C}}_{\mathbf{x}|\mathbf{z}} =\ist \mathbf{C}_{\mathbf{x}} 
  - \tilde{\mathbf{K}} (\tilde{\mathbf{C}}_{\mathbf{y}} \rmv+\rmv \mathbf{C}_{\mathbf{n}}) \tilde{\mathbf{K}}^{\text{T}} \rmv, 
\label{eq:MessUpdate_t}
\vspace*{.7mm}
\ee
with $\tilde{\mathbf{K}} = \tilde{\mathbf{C}}_{\mathbf{x}\mathbf{y}}(\tilde{\mathbf{C}}_{\mathbf{y}} \rmv+\rmv \mathbf{C}_{\mathbf{n}})^{-1}\rmv$.
We thus obtain the following approximate SP implementation of \eqref{eq:BayesianRule}.

\emph{Step 1}:\, SPs and weights $\big\{ \rmv\big( \mathbf{x}^{(j)}\rmv,w_{\text{m}}^{(j)}\rmv,w_{\text{c}}^{(j)} \big)\rmv \big\}_{j=0}^{2J}\rmv$ 
are calculated from $\bm{\mu}_{\mathbf{x}}$ and $\mathbf{C}_{\mathbf{x}}$ \cite{haykin2001ch7}.

\emph{Step 2}:\, The transformed SPs $\mathbf{y}^{(j)} \rmv= H\big(\mathbf{x}^{(j)}\big)$, $j \rmv\in\rmv \{0,\ldots,$\linebreak %%%%%
$2J\}$ are calculated.

\emph{Step 3}:\, From $\big\{ \rmv\big( \mathbf{x}^{(j)}\rmv, \mathbf{y}^{(j)}\rmv, w_{\text{m}}^{(j)}\rmv, w_{\text{c}}^{(j)} \big)\rmv \big\}_{j=0}^{2J}$, 
the means and covariances $\tilde{\bm{\mu}}_{\mathbf{y}}$, $\tilde{\mathbf{C}}_{\mathbf{y}}$, and $\tilde{\mathbf{C}}_{\mathbf{x}\mathbf{y}}$ 
in \eqref{eq:MUandCOV_my}--\eqref{eq:MUandCOV_cxy} and, in turn, 
$\tilde{\bm{\mu}}_{\mathbf{x}|\mathbf{z}}$ and $\tilde{\mathbf{C}}_{\mathbf{x}|\mathbf{z}}$ in 
\eqref{eq:MessUpdate_t} are calculated.

This algorithm can be extended to nonadditive noise \cite{haykin2001ch7}.

\vspace{-2mm}

%%%%%%%%%%%%%%%%%%%%%%%%%%%%%%%%%%%%%%
\section{Belief Propagation}
\label{sec:BP}
%%%%%%%%%%%%%%%%%%%%%%%%%%%%%%%%%%%%%%

We will next describe our system model and review BP. We consider $K$ states $\mathbf{x}_{k}$, $k \rmv\in\rmv \{1,\ldots,K\}$ and 
observations $\mathbf{z}_{k,l}$ that involve pairs of states $\mathbf{x}_{k}, \mathbf{x}_{l}$ according to 
\begin{equation}
\label{eq:messmodCL}
\mathbf{z}_{k,l} \,=\, G(\mathbf{x}_{k},\mathbf{x}_{l}) + \mathbf{n}_{k,l} \,, \quad (k,l) \rmv\in\rmv \mathcal{E} \,.
\end{equation}
Here, the set $\mathcal{E} \rmv\subseteq\rmv \{1,\ldots,K\}^2$ is symmetric, i.e., $(k,l) \rmv\in\rmv \mathcal{E}$ implies $(l,k) \rmv\in\rmv \mathcal{E}$; 
$G(\cdot,\cdot)$ is a generally nonlinear symmetric function, i.e., $G(\mathbf{x}_{k},\mathbf{x}_{l}) \rmv=\rmv G(\mathbf{x}_{l},\mathbf{x}_{k})$;
and $\mathbf{n}_{k,l} \rmv=\rmv \mathbf{n}_{l,k}$ is zero-mean with known covariance matrix $\mathbf{C}_{\mathbf{n}_{k,l}}$. Note that 
$\mathbf{z}_{k,l} \rmv= \mathbf{z}_{l,k}$. (Methods for enforcing symmetry and a modified BP scheme that does not require it will be discussed 
presently.) We assume that $\mathbf{n}_{k,l}$ is independent of all $\mathbf{x}_k$, that $\mathbf{n}_{k,l}$ and $\mathbf{n}_{k'\!,l'}$ are
independent unless 
\pagebreak %%%%%%
$(k,l) \rmv=\rmv (k'\!,l')$ or $(k,l) \rmv=\rmv (l'\!,k')$, and that all $\mathbf{x}_{k}$ are \emph{a priori} independent. 
The BP algorithm---as well as the SPBP algorithm presented in Section \ref{sec:SPBP}---can be easily extended to more general system models, i.e.,
it is not limited to additive noise and ``pairwise'' observations.

In what follows, $\mathbf{x} \triangleq (\mathbf{x}_1^{\text{T}} \cdots\ist \mathbf{x}_K^{\text{T}})^{\text{T}}\rmv$; similarly,
$\mathbf{z}$ is defined by stacking all $\mathbf{z}_{k,l}$ in arbitrary order. Because of \eqref{eq:BayesianRule} and our assumptions, 
the joint posterior pdf $f(\mathbf{x}|\mathbf{z})$ factorizes as 
\begin{equation}
\label{eq:factorization}
f(\mathbf{x}|\mathbf{z}) \,\propto\ist \prod_{k=1}^K f(\mathbf{x}_{k}) 
\!\!\!\!\prod_{\begin{array}{c} \rule{1mm}{0mm}\\[-5.2mm]{\scriptstyle (k'\!\rmv,l) \in \mathcal{E}}\\[-1.3mm]{\scriptstyle k'\rmv > l} \end{array}} \!\!\!\!\! 
  f(\mathbf{z}_{k'\!\rmv,l}|\mathbf{x}_{k'}\rmv,\mathbf{x}_l) \,.
 \vspace{-1mm}
\end{equation}
Bayesian estimation of the states $\mathbf{x}_k$ relies on the marginal posterior pdfs $f(\mathbf{x}_k|\mathbf{z})$. Whereas direct marginalization of 
$f(\mathbf{x}|\mathbf{z})$ in \eqref{eq:factorization} is usually infeasible, approximate marginal posterior pdfs (``beliefs'') 
$b(\mathbf{x}_k) \approx f(\mathbf{x}_k|\mathbf{z})$ can be computed by executing iterative BP message passing on the factor graph corresponding to 
\eqref{eq:factorization} \cite{loeliger}. Let the ``neighbor'' set $\mathcal{N}_k \rmv= \{ l_1, l_2, \dots, l_{|\mathcal{N}_k|} \}$ of variable node 
$k \rmv\in\rmv \{1,\ldots,K\}$ comprise all $l \rmv\in\rmv \{1,\ldots,K\} \!\setminus\! \{k\}$ such that $(k,l) \rmv\in\rmv \mathcal{E}$.
Then, at iteration $p \rmv\ge\rmv 1$, the belief of variable node $k$ is obtained as \cite{loeliger}
\begin{equation}
\label{eq:BPrules_b}
b^{(p)}(\mathbf{x}_k) \,\propto\, f(\mathbf{x}_{k}) \rmv\prod_{l \in \mathcal{N}_k} \!m^{(p)}_{l\rightarrow k}(\mathbf{x}_k) \,,
\vspace{-3mm}
\end{equation}
where 
\be
\hspace*{-2mm}m^{(p)}_{l\rightarrow k}(\mathbf{x}_k) \ist=\rmv \int \rmv\rmv f(\mathbf{z}_{k,l}|\mathbf{x}_k,\mathbf{x}_l) \, n^{(p-1)}_{l\rightarrow k}(\mathbf{x}_l) \, d\mathbf{x}_l \,, 
  \quad\!\rmv l \rmv\in\rmv \mathcal{N}_k \,,  
\label{eq:BPrules_m}
\ee
with
\be  
n^{(p)}_{l\rightarrow k}(\mathbf{x}_l) \,=\, f(\mathbf{x}_{l}) \rmv\!\prod_{k' \in \mathcal{N}_l \backslash \{k\}} \!\! m^{(p)}_{k'\rightarrow l}(\mathbf{x}_l) \,, 
  \quad\! l \rmv\in\rmv \mathcal{N}_k \,. 
\label{eq:BPrules_n}
\vspace{-1mm}
\ee
This recursion is initialized by setting $n^{(0)}_{l\rightarrow k}(\mathbf{x}_l)$ equal to the prior pdf of $\mathbf{x}_{l}$. 

In a decentralized scenario where the variable nodes in the factor graph are simultaneously sensor nodes, the symmetry condition 
$\mathbf{z}_{k,l} \rmv=\rmv \mathbf{z}_{l,k}$ is usually not satisfied. It can be enforced by considering the averaged measurement 
$(\mathbf{z}_{k,l} + \mathbf{z}_{l,k})/2$ or the stacked measurement $(\mathbf{z}_{k,l}^{\text{T}} \;\, \mathbf{z}_{l,k}^{\text{T}})^{\text{T}}\rmv$;
this requires communication between the sensor nodes $k$ and $l$. Alternatively, the condition $\mathbf{z}_{k,l} \rmv=\rmv \mathbf{z}_{l,k}$ 
can be avoided by using the SPAWN message passing scheme \cite{wymeersch}. SPAWN differs from the standard BP operations 
\eqref{eq:BPrules_b}--\eqref{eq:BPrules_n} in that $n^{(p)}_{l\rightarrow k}(\mathbf{x}_l) \rmv=\rmv b^{(p)}(\mathbf{x}_l)$ for all $l \rmv\in\rmv \mathcal{N}_k$. 
Consequently, \eqref{eq:BPrules_m} is replaced by
\[
\hspace*{-2mm}m^{(p)}_{l\rightarrow k}(\mathbf{x}_k) \ist=\rmv \int \rmv\rmv f(\mathbf{z}_{k,l}|\mathbf{x}_k,\mathbf{x}_l) \, b^{(p-1)}(\mathbf{x}_l) \, d\mathbf{x}_l \,, 
  \quad\! l \rmv\in\rmv \mathcal{N}_k \,,
\]
and \eqref{eq:BPrules_n} need not be calculated.

\vspace{-1mm}

%%%%%%%%%%%%%%%%%%%%%%%%%%%%%%%%%%%%%%
\section{Sigma Point Belief Propagation}
\label{sec:SPBP}
%%%%%%%%%%%%%%%%%%%%%%%%%%%%%%%%%%%%%%

\vspace{.7mm}

The proposed SPBP algorithm performs a low-complexity, approximate calculation of $b^{(p)}(\mathbf{x}_k)$ in \eqref{eq:BPrules_b} based on SPs. 
The key idea is to reformulate the BP operations in higher-dimensional spaces that are defined by the ``composite'' vectors 
$\bar{\mathbf{x}}_k \rmv\triangleq\rmv \big( \mathbf{x}_k^{\text{T}} \; \mathbf{x}_{l_1}^{\text{T}} \,\ist \mathbf{x}_{l_2}^{\text{T}} \rmv\cdots\ist 
\mathbf{x}_{l_{|\mathcal{N}_k|}}^{\text{T}} \big)^{\rmv\text{T}}\rmv$ ($\mathbf{x}_k$ and its neighbor states) and 
$\bar{\mathbf{z}}_k \rmv\triangleq\rmv \big( \mathbf{z}_{k,l_1}^{\text{T}} \; \mathbf{z}_{k,l_2}^{\text{T}} 
\rmv\cdots\ist \mathbf{z}_{k,l_{|\mathcal{N}_k|}}^{\text{T}} \big)^{\rmv\text{T}}\rmv$ (all observations involving $\mathbf{x}_k$). 
The dimension of $\bar{\mathbf{x}}_k$ is $\bar{J}_k \triangleq J_k + \sum_{l=1}^{|\mathcal{N}_k|} J_l$, where $J_k$ denotes the dimension of $\mathbf{x}_k$. 
Let $\bar{\mathbf{x}}_k^{\sim k}$ (resp.\ $\bar{\mathbf{x}}^{\sim k,l}_{k}$) denote $\bar{\mathbf{x}}_k$ with the subvector $\mathbf{x}_k$ 
(resp.\ the subvectors $\mathbf{x}_k$ and $\mathbf{x}_l$) removed, and let $\bar{\mathbf{z}}_k^{\sim l}$ denote $\bar{\mathbf{z}}_k$ 
with the subvector $\mathbf{z}_{k,l}$ removed. By inserting \eqref{eq:BPrules_m} into \eqref{eq:BPrules_b} and \eqref{eq:BPrules_n}, we obtain
\begin{align}
b^{(p)}(\mathbf{x}_k) &\,\propto \int 
  \rmv\rmv f(\bar{\mathbf{z}}_k|\bar{\mathbf{x}}_k) \ist f^{(p-1)}(\bar{\mathbf{x}}_k) \, d \bar{\mathbf{x}}_k^{\sim k} \label{eq:ref_marg}\\[.5mm]
\hspace*{-8mm}n^{(p)}_{l\rightarrow k}(\mathbf{x}_l) &\,= \int \rmv\rmv f(\bar{\mathbf{z}}_l^{\sim k}|\bar{\mathbf{x}}_l^{\sim k}) 
  \ist f^{(p-1)}(\bar{\mathbf{x}}_l^{\sim k}) \, d \bar{\mathbf{x}}^{\sim l,k}_{l} ,  \quad\!\! l \rmv\in\rmv \mathcal{N}_k \,,\nonumber \\[-2mm] 
& \label{eq:ext_mess}\\[-9mm]
\nonumber
\end{align}
where 
\vspace{-1.5mm}
\begin{align*}
f(\bar{\mathbf{z}}_{k}|\bar{\mathbf{x}}_k) &\ist= \prod_{l \in \mathcal{N}_k} \! f(\mathbf{z}_{k,l}|\mathbf{x}_k,\mathbf{x}_l)\\
f^{(p-1)}(\bar{\mathbf{x}}_k) &\ist\propto\ist f(\mathbf{x}_{k}) \rmv \prod_{l \in \mathcal{N}_k} \! n^{(p-1)}_{l\rightarrow k}(\mathbf{x}_l)\\ 
f(\bar{\mathbf{z}}_l^{\sim k}|\bar{\mathbf{x}}_l^{\sim k}) &\ist= \prod_{k' \in \mathcal{N}_l \rmv\backslash \{k\}} \!\rmv f(\mathbf{z}_{l,k'}|\mathbf{x}_l,\mathbf{x}_{k'})\\ 
f^{(p-1)}(\bar{\mathbf{x}}_l^{\sim k}) &\ist\propto\ist f(\mathbf{x}_{l}) \rmv \prod_{k' \rmv\in \mathcal{N}_l \rmv\backslash \{k\}} \!\rmv n^{(p-1)}_{k'\rmv\rightarrow l}(\mathbf{x}_{k'}) \,. 
\end{align*}
(If SPAWN is used, $f^{(p-1)}(\bar{\mathbf{x}}_k) \propto f(\mathbf{x}_{k}) \ist \prod_{l \in \mathcal{N}_k} \rmv b^{(p-1)}(\mathbf{x}_l)$, 
and thus $n^{(p)}_{l\rightarrow k}(\mathbf{x}_l)$ need not be calculated.) Note that $f(\bar{\mathbf{z}}_k|\bar{\mathbf{x}}_k)$ and 
$f^{(p-1)}(\bar{\mathbf{x}}_k)$ are, respectively, the likelihood function and ``iterated prior pdf'' for the composite observation 
\vspace{-1mm}
model
\[
\bar{\mathbf{z}}_k \ist=\ist \bar{\mathbf{y}}_k  +\ist \bar{\mathbf{n}}_k \,, \quad \text{with} \;\, \bar{\mathbf{y}}_k  = H(\bar{\mathbf{x}}_k) \,,
\]
where $H(\bar{\mathbf{x}}_k) \triangleq \big( \ist \big( G(\mathbf{x}_{k},\mathbf{x}_{l_{1}}) \big)^{\rmv\text{T}} 
  \cdots\ist \big( G(\mathbf{x}_{k},\mathbf{x}_{l_{|\mathcal{N}_k|}}) \big)^{\rmv\text{T}} \ist \big)^{\rmv\text{T}}$ 
and $\bar{\mathbf{n}}_k \triangleq \big( \mathbf{n}_{k,l_1}^{\text{T}} \rmv\cdots\ist \mathbf{n}_{k,l_{|\mathcal{N}_k|}}^{\text{T}} \big)^{\rmv\text{T}}$.

To develop an SP-based approximate calculation of $b^{(p)}(\mathbf{x}_k)$, we first note that \eqref{eq:ref_marg} is a marginalization, i.e.,
\be
b^{(p)}(\mathbf{x}_k) \,= \int \! b^{(p)}(\bar{\mathbf{x}}_k) \, d \bar{\mathbf{x}}_k^{\sim k} \ist,
\label{eq:b_marg}
\vspace{-3mm}
\ee
with 
\vspace{1mm}
\be
b^{(p)}(\bar{\mathbf{x}}_k) \,\propto\, f(\bar{\mathbf{z}}_k|\bar{\mathbf{x}}_k) \, f^{(p-1)}(\bar{\mathbf{x}}_k) \,.
\label{eq:b_b}
\ee
Because the expression \eqref{eq:b_b} of the ``composite belief'' $b^{(p)}(\bar{\mathbf{x}}_k)$
is analogous to \eqref{eq:BayesianRule}, we can obtain an approximate SP representation of $b^{(p)}(\bar{\mathbf{x}}_k)$ in a similar way as 
we obtained an approximate SP representation of $f(\mathbf{x}|\mathbf{z})$ in Section \ref{sec:SPbasics}. We first define a mean vector
and a covariance matrix corresponding to the ``composite prior'' 
\vspace{-.3mm}
$f^{(p-1)}(\bar{\mathbf{x}}_k) \propto  f(\mathbf{x}_{k}) \ist \prod_{l \in \mathcal{N}_k} \! n^{(p-1)}_{l\rightarrow k}(\mathbf{x}_l)$:
\begin{align}
\hspace{-1mm}\bm{\mu}^{(p-1)}_{\bar{\mathbf{x}}_k} &\ist\triangleq\, \Big( \ist \bm{\mu}_{\mathbf{x}_k}^{\text{T}} \;\, \bm{\mu}^{(p-1)\ist\text{T}}_{l_{1}\rightarrow k} \;\,
  \bm{\mu}^{(p-1)\ist\text{T}}_{l_{2}\rightarrow k} \ist\cdots\ist \bm{\mu}^{(p-1)\ist\text{T}}_{l_{|\mathcal{N}_k|}\rightarrow k} \Big)^{\text{T}} \label{comp_prior}\\[1mm]
\hspace{-1mm}\mathbf{C}^{(p-1)}_{\bar{\mathbf{x}}_k} &\ist\triangleq\, \mathrm{diag} \Big\{ \mathbf{C}_{\mathbf{x}_k}, \mathbf{C}^{(p-1)}_{l_{1}\rightarrow k},
  \mathbf{C}^{(p-1)}_{l_{2}\rightarrow k}, \ldots, \mathbf{C}^{(p-1)}_{l_{|\mathcal{N}_k|}\rightarrow k} \Big\} \ist. \label{comp_prior_1}\\[-6mm]
\nonumber
\end{align}
Here, we interpreted $\prod_{l \in \mathcal{N}_k} \! n^{(p-1)}_{l\rightarrow k}(\mathbf{x}_l)$ as the product of the pdfs of statistically independent 
random variables (up to a normalization); furthermore, $\bm{\mu}_{\mathbf{x}_k}$ and $\mathbf{C}_{\mathbf{x}_k}$ are the mean and covariance matrix of 
the prior $f(\mathbf{x}_{k})$; $\bm{\mu}^{(p-1)}_{l_{i}\rightarrow k}$ and $\mathbf{C}^{(p-1)}_{l_{i}\rightarrow k}$ are the mean and covariance matrix of 
$n^{(p-1)}_{l_i\rightarrow k}(\mathbf{x}_{l_i})$; and $\mathrm{diag} \{\cdot\}$ denotes the block diagonal matrix whose diagonal blocks are the listed matrices.
(For SPAWN, $\bm{\mu}^{(p-1)}_{l_{i}\rightarrow k}$ and $\mathbf{C}^{(p-1)}_{l_{i}\rightarrow k}$ are the mean and covariance matrix of $b^{(p-1)}(\mathbf{x}_{l_i})$.)
The following steps are now performed for each $k \rmv\in\rmv \{1,\ldots,K\}$ (note that the first three steps are analogous to those in Section \ref{sec:SPbasics}).

\emph{Step 1}:\, SPs and weights $\big\{ \rmv\big( \bar{\mathbf{x}}_k^{(j)}\rmv, w_{\text{m}}^{(j)}\rmv, w_{\text{c}}^{(j)} \big)\rmv \big\}_{j=0}^{2\bar{J}_k}\rmv$ 
corresponding to $f^{(p-1)}(\bar{\mathbf{x}}_k)$ are calculated 
\pagebreak %%%%%%
from $\bm{\mu}^{(p-1)}_{\bar{\mathbf{x}}_k}$ and $\mathbf{C}^{(p-1)}_{\bar{\mathbf{x}}_k}$ \cite{haykin2001ch7}.  
(Note that the dimension and number of the SPs depend on the number of neighbors $|\mathcal{N}_k|$, and thus
the tuning parameters that adjust the spread of the SPs should be adapted to $|\mathcal{N}_k|$.)

\emph{Step 2}:\, The transformed SPs $\bar{\mathbf{y}}_k^{(j)} \rmv= H\big(\bar{\mathbf{x}}_k^{(j)}\big)$, $j \rmv\in\rmv \{0,\ldots,$\linebreak %%%%%
$2\bar{J}_k\}$ are calculated.

\emph{Step 3}:\, From $\big\{ \rmv\big( \bar{\mathbf{x}}_k^{(j)}\rmv, \bar{\mathbf{y}}_k^{(j)}\rmv, w_{\text{m}}^{(j)}\rmv, w_{\text{c}}^{(j)} \big)\rmv \big\}_{j=0}^{2\bar{J}_k}$, 
the means and covariances $\tilde{\bm{\mu}}_{\bar{\mathbf{y}}_k}^{(p)}$, $\tilde{\mathbf{C}}_{\bar{\mathbf{y}}_k}^{(p)}$, and  
$\tilde{\mathbf{C}}_{\bar{\mathbf{x}}_k\bar{\mathbf{y}}_k}^{(p)}$ are calculated as in \eqref{eq:MUandCOV_my}--\eqref{eq:MUandCOV_cxy}. 
Subsequently, $\tilde{\bm{\mu}}^{(p)}_{b(\bar{\mathbf{x}}_k)}$ and $\tilde{\mathbf{C}}^{(p)}_{b(\bar{\mathbf{x}}_k)}$ (the SP approximations of the mean 
and covariance matrix of $b^{(p)}(\bar{\mathbf{x}}_k)$) are calculated as in \eqref{eq:MessUpdate_t}, using $\tilde{\bm{\mu}}_{\bar{\mathbf{y}}_k}^{(p)}$, 
$\tilde{\mathbf{C}}_{\bar{\mathbf{y}}_k}^{(p)}$, and $\tilde{\mathbf{C}}_{\bar{\mathbf{x}}_k\bar{\mathbf{y}}_k}^{(p)}$ 
instead of $\tilde{\bm{\mu}}_{\mathbf{y}}$, $\tilde{\mathbf{C}}_{\mathbf{y}}$, and $\tilde{\mathbf{C}}_{\mathbf{xy}}$, respectively. 

\emph{Step 4}:\, From $\tilde{\bm{\mu}}^{(p)}_{b(\bar{\mathbf{x}}_k)}$ and $\tilde{\mathbf{C}}^{(p)}_{b(\bar{\mathbf{x}}_k)}$, 
the elements related to $\mathbf{x}_{k}$\linebreak %%%%%% 
are extracted (this corresponds to the marginalization \eqref{eq:b_marg}). More specifically, the approximate mean $\tilde{\bm{\mu}}^{(p)}_{b(\mathbf{x}_k)}$ 
and covariance matrix $\tilde{\mathbf{C}}^{(p)}_{b(\mathbf{x}_k)}$ of the ``marginal belief'' $b^{(p)}(\mathbf{x}_k)$ are given by the first $J_k$ elements of 
$\tilde{\bm{\mu}}^{(p)}_{b(\bar{\mathbf{x}}_k)}$ and the upper-left $J_k \!\times\! J_k$ submatrix of $\tilde{\mathbf{C}}^{(p)}_{b(\bar{\mathbf{x}}_k)}$, respectively 
(cf.\ the stacked structure of $\bm{\mu}^{(p-1)}_{\bar{\mathbf{x}}_k}$ in \eqref{comp_prior} and the block structure of $\mathbf{C}^{(p-1)}_{\bar{\mathbf{x}}_k}$ 
in \eqref{comp_prior_1}).

An SP-based approximate calculation of the messages $n^{(p)}_{l\rightarrow k}(\mathbf{x}_l)$, $l \rmv\in\rmv \mathcal{N}_k$ in \eqref{eq:ext_mess} 
can be performed in a similar manner, due to the structural analogy of \eqref{eq:ext_mess} to \eqref{eq:ref_marg}. (For SPAWN, 
$n^{(p)}_{l\rightarrow k}(\mathbf{x}_l)$ is not needed.) We note that, as loopy BP in general \cite{wymeersch12}, SPBP typically exhibits convergence 
of the mean but suffers from overconfident covariance matrices.

\vspace{-1mm}

%%%%%%%%%%%%%%%%%%%%%%%%%%%%%%%%%%%%%%
\section{Computation and Communication Requirements}
\label{sec:compcom}
%%%%%%%%%%%%%%%%%%%%%%%%%%%%%%%%%%%%%%

\vspace{.7mm}

Similar to the SP filter \cite{haykin2001ch7}, SPBP requires the computation of the square root of the $\bar{J}_k \rmv\times\rmv \bar{J}_k$ matrices 
$\mathbf{C}^{(p-1)}_{\bar{\mathbf{x}}_k}$ to calculate the SPs $\bar{\mathbf{x}}_k^{(j)}$ in Step 1. This is the most complex part of the SPBP algorithm.
An efficient computation of the matrix square root uses the Cholesky decomposition \cite{press92}, whose complexity is cubic in 
$\bar{J}_k = J_k + \sum_{l=1}^{|\mathcal{N}_k|} J_l$. Thus, the complexity of SPBP is cubic in $|\mathcal{N}_k|$ and, also, in the number of SPs
(which is $2\bar{J}_k + 1$). The complexity of NBP is linear in $|\mathcal{N}_k|$ and quadratic in the number of particles \cite{lien}. 
However, the number of particles in NBP is usually much higher than the number of SPs in SPBP. Moreover, the quadratic and cubic complexity terms 
of the Cholesky decomposition are rather small (about $\bar{J}_k^3/6$ multiplications, $\bar{J}_k^2/2$ divisions, and $\bar{J}_k$ square root operations 
are used \cite{press92}). Therefore, in many applications, SPBP is significantly less complex than NBP.

SPBP is especially advantageous in decentralized signal processing applications where each variable node in the factor graph corresponds to a 
sensor node in a wireless sensor network. Because $b^{(p)}(\mathbf{x}_k)$ and $n^{(p)}_{k\rightarrow l}(\mathbf{x}_k)$ are represented 
by a mean vector and a covariance matrix, at most $J_k + \frac{J_k(J_k+1)}{2} =$\linebreak %%%%%%
$\frac{J_k(J_k+3)}{2}$ real values per message passing iteration 
$p \rmv\in\rmv \{1,\ldots,$\linebreak %%%%%%
$P\}$ have to be transmitted from sensor $k$ to neighboring sensor nodes, rather than hundreds or thousands of particles in NBP.
More specifically, at message passing iteration $p$, sensor $k$ receives  $\bm{\mu}^{(p-1)}_{l\rightarrow k}$ and $\mathbf{C}^{(p-1)}_{l\rightarrow k}$ 
from all $l \rmv\in\rmv \mathcal{N}_k$ (this is needed to calculate $\bm{\mu}^{(p-1)}_{\bar{\mathbf{x}}_k}$ and $\mathbf{C}^{(p-1)}_{\bar{\mathbf{x}}_k}$,
see \eqref{comp_prior} and \eqref{comp_prior_1}), and it broadcasts $\bm{\mu}^{(p-1)}_{k\rightarrow l}$ and $\mathbf{C}^{(p-1)}_{k\rightarrow l}$ 
to all $l \rmv\in\rmv \mathcal{N}_k$. These communications are a precondition for Step 1 of the SPBP algorithm. 
If the measurement model in \eqref{eq:messmodCL} involves only substates $\bm{\lambda}_{k}$ of the states $\mathbf{x}_k$, only the mean 
and covariance matrix corresponding to $\bm{\lambda}_{k}$ have to be transmitted. (In NBP, similarly, only subparticles corresponding to 
$\bm{\lambda}_{k}$ have to be transmitted.)

\vspace{-1.5mm}

%%%%%%%%%%%%%%%%%%%%%%%%%%%%%%%%%%%%%%
\section{Simulation Results}
\label{sec:simres}
%%%%%%%%%%%%%%%%%%%%%%%%%%%%%%%%%%%%%%

We simulated\footnote{The %%%%%%%
simulation source files and further information about the simulation setting are available online at 
http://www.nt.tuwien.ac.at/about-us/staff/florian-meyer/.} %%%%%%%
a decentralized, cooperative, dynamic self-localization scenario \cite{wymeersch} using a network of $K\!\rmv=\!5$ sensors, of 
which three are mobile and two are anchors, i.e., static sensors with perfect location information. The state of mobile sensor $k \rmv\in\rmv \{1,2,3\}$ at time  
$i \rmv\in\rmv \{0,1,\ldots,50\}$ consists of the location $\bm{\lambda}_{k,i} \rmv\triangleq\rmv (x_{1,k,i}\,\iist x_{2,k,i} )^\text{T}\rmv$
and the velocity, i.e., $\mathbf{x}_{k,i} \rmv\triangleq\rmv (x_{1,k,i}\,\iist x_{2,k,i} \,\iist v_{1,k,i} \,\iist v_{2,k,i})^\text{T}\rmv$.
Each mobile sensor moves within a field of size 50$\ist\times\ist$50, performs distance measurements relative to all other sensors, 
communicates the mean and covariance matrix of its current location to all other sensors, and estimates its own state. 
We assume that each mobile sensor is able to associate its measurements with the individual sensors. Each anchor sensor $k \!\in\! \{4,5\}$ 
communicates its own (true) location $\bar{\bm{\lambda}}_{k}$. The distance measurement of mobile sensor $k \rmv\in \{1,2,3\}$ relative to 
sensor $l$ at time $i$
\vspace{-.5mm}
is (cf.\ \eqref{eq:messmodCL}) 
\[
z_{k,l,i} \ist= \begin{cases}
\| \bm{\lambda}_{k,i} - \bm{\lambda}_{l,i} \| + n_{k,l,i}\,, & l \rmv\in\rmv \{1,2,3\} \!\setminus \! \{k\}\\[.4mm]
\| \bm{\lambda}_{k,i} - \bar{\bm{\lambda}}_{l} \| + n_{k,l,i}\,, & l \rmv\in\rmv \{4,5\} \,,
\end{cases}
\vspace{-.5mm}
\]
where $n_{k,l,i}$ is zero-mean Gaussian measurement noise with variance $\sigma_n^2 \!=\! 1$.

The states of the mobile sensors evolve independently according to $\mathbf{x}_{k,i} = \mathbf{G}\mathbf{x}_{k,i-1} + \mathbf{W}\mathbf{u}_{k,i}$ \cite{rong}.
Here, the matrices $\mathbf{G} \!\in\! \mathbb{R}^{4\times 4}$ and $\mathbf{W} \rmv\!\in\! \mathbb{R}^{4\times 2}$ are chosen as in \cite{kotecha03} 
and the driving noise vectors $\mathbf{u}_{k,i} \!\in\! \mathbb{R}^2$ are Gaussian, i.e., $\mathbf{u}_{k,i} \!\sim\! \mathcal{N}(\mathbf{0},\sigma_u^2\mathbf{I})$, 
with variance $\sigma_{u}^2 \!=\! 10^{-4}$; furthermore, $\mathbf{u}_{k,i}$ and $\mathbf{u}_{k'\!,i'}$ are independent unless $(k,i) \!=\! (k'\!,i')$. 
In the generation of the state sequences, this recursive evolution of the $\mathbf{x}_{k,i}$ was initialized with 
$\mathbf{x}_{1,0} \rmv=\rmv (0 \,\,\ist 0 \,\,\ist 0.2 \,\,\ist 1)^\text{T}\rmv$, $\mathbf{x}_{2,0} \rmv= (25 \,\,\ist 50 \,\,\ist 0.5 \,\, {-0.8})^\text{T}\rmv$, and
$\mathbf{x}_{3,0} \rmv=\rmv (50 \,\,\ist 0 \,\, {-1} \,\,\ist 0.4)^\text{T}\rmv$. The anchor sensors are located at $\bar{\bm{\lambda}}_{4} \rmv=\rmv (0 \,\,\ist 25)^\text{T}\rmv$ 
and $\bar{\bm{\lambda}}_{5} \rmv= (50 \,\,\ist 25)^\text{T}$ for all $i$. In the simulation of the various self-localization algorithms, for the mobile sensors, 
we used the initial prior pdf $f(\mathbf{x}_{k,0}) \rmv =\rmv \mathcal{N}(\bm{\mu}_{k,0},\mathbf{C}_{k,0} )$. Here, 
$\mathbf{C}_{k,0} \rmv=\rmv \mathrm{diag}\ist\{1, 1, 0.01, 0.01\}$ represents the uncertainy in knowing $\mathbf{x}_{k,0}$, and $\bm{\mu}_{k,0}$ 
is a random hyperparameter that was randomly sampled (for each simulation run) from $\mathcal{N}(\mathbf{x}_{k,0},\mathbf{C}_{k,0})$.
For the anchor sensors, the true locations were used. The number of message passing iterations $p$ at each time $i$ was set to $P \rmv=\rmv 2$.

\begin{figure}[t!]
\centering
\psfrag{s01}[t][t][0.7]{\color[rgb]{0,0,0}\setlength{\tabcolsep}{0pt}\begin{tabular}{c}\raisebox{-2mm}{$i$}\end{tabular}}
\psfrag{s02}[b][b][0.5]{\color[rgb]{0,0,0}\setlength{\tabcolsep}{0pt}\begin{tabular}{c}\vspace{0mm}{\Large RMSE}\end{tabular}}
\psfrag{s05}[l][l][0.6]{\color[rgb]{0,0,0}SPBP}
\psfrag{s06}[l][l][0.6]{\color[rgb]{0,0,0}NBP-2\ist\ist (250)}
\psfrag{s07}[l][l][0.6]{\color[rgb]{0,0,0}NBP-1\ist\ist (250)}
\psfrag{s08}[l][l][0.6]{\color[rgb]{0,0,0}NBP-2\ist\ist (500)}
\psfrag{s09}[l][l][0.6]{\color[rgb]{0,0,0}NBP-1\ist (500)}
\psfrag{s10}[l][l][0.6]{\color[rgb]{0,0,0}NBP-2\ist\ist (1000)}
\psfrag{s11}[l][l][0.6]{\color[rgb]{0,0,0}NBP-1\ist (1000)}
\psfrag{s12}[l][l][0.6]{\color[rgb]{0,0,0}SPBP}
\psfrag{s14}[][]{\color[rgb]{0,0,0}\setlength{\tabcolsep}{0pt}\begin{tabular}{c} \end{tabular}}
\psfrag{s15}[][]{\color[rgb]{0,0,0}\setlength{\tabcolsep}{0pt}\begin{tabular}{c} \end{tabular}}
\psfrag{x01}[t][t][0.65]{$0$}
\psfrag{x02}[t][t][0.65]{$5$}
\psfrag{x03}[t][t][0.65]{$10$}
\psfrag{x04}[t][t][0.65]{$15$}
\psfrag{x05}[t][t][0.65]{$20$}
\psfrag{x06}[t][t][0.65]{$25$}
\psfrag{x07}[t][t][0.65]{$30$}
\psfrag{x08}[t][t][0.65]{$35$}
\psfrag{x09}[t][t][0.65]{$40$}
\psfrag{x10}[t][t][0.65]{$45$}
\psfrag{x11}[t][t][0.65]{$50$}

\psfrag{v01}[r][r][0.65]{$0.4$}
\psfrag{v02}[r][r][0.65]{$0.6$}
\psfrag{v03}[r][r][0.65]{$0.8$}
\psfrag{v04}[r][r][0.65]{$1.0$}
\psfrag{v05}[r][r][0.65]{$1.2$}
\psfrag{v06}[r][r][0.65]{$1.4$}
\psfrag{v07}[r][r][0.65]{$1.6$}
\psfrag{v08}[r][r][0.65]{$1.8$}
\hspace*{-1mm}\includegraphics[height=35mm, width=60mm]{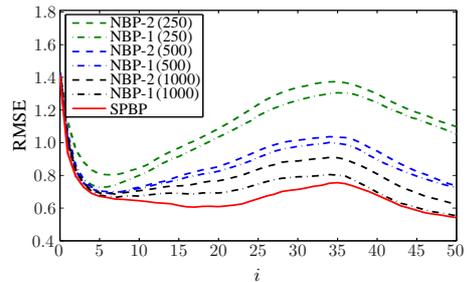}
\vspace{.5mm}
\caption{Root-mean-square location and velocity error (RMSE) of the simulated self-localiza\-tion algorithms versus time $i$.}
\label{fig:plot}
\vspace{-1mm}
\end{figure}

We compare the proposed SPBP algorithm (using 25 SPs) with two NBP methods for cooperative self-localization, referred to as NBP-1 and NBP-2.
NBP-1 \cite{lien} is an extension of the method in \cite{ihler} to moving sensors. NBP-2 differs from NBP-1 in that it performs the message multiplication 
\eqref{eq:BPrules_b} using Monte Carlo integration instead of Gaussian kernels \cite{savic13}. All three methods are based on SPAWN. The NBP methods 
use 250, 500, or 1000 particles. In NBP-1, the bandwidth of the Gaussian kernels was equal to the measurement noise variance $\sigma_n^2 \!=\! 1$ \cite{ihler}. 
Fig.\ \ref{fig:plot} shows the simulated root-mean-square location and velocity error of the various methods for $i = 1,\ldots,50$. 
This error was determined by averaging over the three mobile sensors and 1000 simulation runs. It is seen that, for the considered simulation parameters, SPBP 
outperforms the two NBP methods. We note, however, that NBP would outperform SPBP if the number of particles in NBP was further increased. Also, 
we expect performance advantages of NBP over SPBP in problems with stronger nonlinearities.

The average runtime of our SPBP implementation on an Intel Xeon X5650 CPU, for all 50 time steps of one simulation run, was $0.61\ist$s.
The average runtime of NBP-1 was $1.53\ist$s, $5.16\ist$s, and $19.57$s (for 250, 500, and 1000 particles, respectively), that of NBP-2 was 
$2.01\ist$s, $7.27\ist$s, and $28.10\ist$s. Thus, in this scenario, SPBP is less complex than the two NBP methods. 

With SPBP, since our measurement model involves only the two-dimensional location $\bm{\lambda}_{k,i}$,
each mobile sensor broadcasts the mean vector and covariance matrix of $b^{(p)}(\bm{\lambda}_{k,i}) = \int \!\int b^{(p)}(\mathbf{x}_{k,i}) \,d v_{1,k,i} \ist d v_{2,k,i}$ at each message passing iteration $p$, corresponding to $2+3=5$ real values. By contrast, for the NBP methods with 250, 500, and 1000 particles, the number of real values
broadcast by each mobile sensor at each message passing iteration is 500, 1000, and 2000, respectively. Thus, SPBP requires significantly less 
communications than the NBP methods. (In all three methods, each anchor sensor broadcasts its location, corresponding to two real values; 
however, this is a preparatory step that is executed only once.)

%% \newpage %%%%%

\vspace{-.5mm}

%%%%%%%%%%%%%%%%%%%%%%%%%%%%%%%%%%%%%%
\section{Conclusion}
%%%%%%%%%%%%%%%%%%%%%%%%%%%%%%%%%%%%%%

We proposed SPBP as a low-complexity approximation of the BP message passing scheme. SPBP extends the SP filter, 
also known as unscented Kalman filter, to nonsequential Bayesian inference for general (loopy) factor graphs. 
Messages and marginal posteriors are represented by mean vectors and covariance matrices, which are calculated using SPs and the unscented 
transformation. Thereby, SPBP avoids both the linearity assumption of Gaussian BP and the typically high complexity of nonparametric (particle-based) BP.
SPBP is especially well suited to certain decentralized inference problems in wireless sensor networks because of its low communication requirements.
In particular, we simulated a decentralized, cooperative, dynamic sensor self-localization scenario and demonstrated significant advantages of SPBP 
over nonparametric BP regarding performance, complexity, and communication requirements.

\renewcommand{\baselinestretch}{1.05}\normalsize\footnotesize
\bibliographystyle{ieeetr_noParentheses}
\bibliography{references}

\end{document}